\documentclass{article}
\usepackage[accepted]{icml2020}
\usepackage[table]{xcolor}
\usepackage{times}
\usepackage{epsfig}
\usepackage{graphicx}
\usepackage{amsmath, mathtools}
\usepackage{amssymb}
\usepackage{pifont}

\usepackage{graphicx}
\usepackage[outdir=./]{epstopdf}
\usepackage{import}
\usepackage{tikz}
\usepackage{pgfplots}

\usepackage{subfig}

\usepackage{enumitem}
\usepackage{booktabs}
\usepackage{tabulary,multirow,overpic}
\usepackage{caption,tabularx,booktabs}
\usepackage{xspace}

\usepackage[]{algorithm2e}
\usetikzlibrary{arrows}
\usetikzlibrary{arrows.meta}
\usetikzlibrary{backgrounds}
\usetikzlibrary{calc}
\usetikzlibrary{chains}
\usetikzlibrary{decorations.markings}
\usetikzlibrary{decorations.pathmorphing}
\usetikzlibrary{fit}
\usetikzlibrary{matrix}
\usetikzlibrary{mindmap}
\usetikzlibrary{pgfplots.colormaps}
\usetikzlibrary{positioning}
\usetikzlibrary{shadows}
\usetikzlibrary{shapes}
\usetikzlibrary{trees}

\definecolor{funkey_bright}{HTML}{FFFFFF}
\definecolor{funkey_lightgrey}{HTML}{cccccc}
\definecolor{funkey_grey}{HTML}{666666}

\definecolor{funkey_color_1}{HTML}{834D9D}
\definecolor{funkey_color_2}{HTML}{F2A431}
\definecolor{funkey_color_3}{HTML}{55B849}

\colorlet{funkey_dark}{purple!10!black}

\colorlet{funkey_highlight}{funkey_color_2}
\colorlet{funkey_textcolor}{funkey_bright}
\colorlet{funkey_bg}{funkey_dark}
\colorlet{funkey_alt_bg}{funkey_lightgrey}

\tikzstyle{patch}=
  [draw=funkey_color_3,thick,inner sep=0]
\tikzstyle{method}=
  [rectangle,rounded corners,draw=funkey_color_2,thick,fill=funkey_color_2!3!white]
\tikzstyle{inset}=
  [rectangle,rounded corners,inner sep=0,clip,draw]
\tikzstyle{arrow}=
  [->,>=latex,very thick]
\tikzstyle{brace}=
  [decorate,decoration={brace,amplitude=5pt,raise=0.5pt},thick]

\newcommand{\getzoomfactor}{%
\pgfgettransformentries{\myxscale}{\@tempa}{\@tempa}{\myyscale}{\@tempa}{\@tempa}
\gdef\zoomfactor{\myxscale}
}

\usetikzlibrary{pgfplots.groupplots}
\usepgfplotslibrary{statistics}
\usepgfplotslibrary{patchplots}

\pgfplotsset{
  errors/.style={
    stack plots=y,
    area style,
    enlarge x limits=false,
    xmajorgrids=true,
    ymajorgrids=true,
    yminorgrids=true,
    legend reversed
  }
}

\pgfplotsset{
  discard if not/.style 2 args={
    x filter/.code={
      \edef\tempa{\thisrow{#1}}
      \edef\tempb{#2}
      \ifx\tempa\tempb
      \else
        \def\pgfmathresult{inf}
      \fi
    }
  }
}

\pgfplotsset{
  discard if not both/.style args={#1 is #2 and #3 is #4}{
    x filter/.code={
      \edef\tempa{\thisrow{#1}}
      \edef\tempb{#2}
      \edef\tempc{\thisrow{#3}}
      \edef\tempd{#4}
      \ifx\tempa\tempb
        \ifx\tempc\tempd
        \else
          \def\pgfmathresult{inf}
        \fi
      \else
        \def\pgfmathresult{inf}
      \fi
    }
  }
}

\pgfplotsset{
  discard if not all three/.style args={#1 is #2 and #3 is #4 and #5 is #6}{
    x filter/.code={
      \edef\tempa{\thisrow{#1}}
      \edef\tempb{#2}
      \edef\tempc{\thisrow{#3}}
      \edef\tempd{#4}
      \edef\tempe{\thisrow{#5}}
      \edef\tempf{#6}
      \ifx\tempa\tempb
        \ifx\tempc\tempd
          \ifx\tempe\tempf
          \else
            \def\pgfmathresult{inf}
          \fi
        \else
          \def\pgfmathresult{inf}
        \fi
      \else
        \def\pgfmathresult{inf}
      \fi
    }
  }
}

\usepackage{grfext}
\PrependGraphicsExtensions*{.jpg,.JPG}

\usepackage[pagebackref=true,breaklinks=true,letterpaper=true,colorlinks,bookmarks=false]{hyperref}
\newlength\savewidth

\makeatletter
\newcolumntype{d}{D{.}{.}{-1} }
\newcolumntype{B}[3]{>{\boldmath\DC@{#1}{#2}{#3} }c<{\DC@end} }
\makeatother

\def\cmark{\ding{51}}

\def\figref#1{Fig.~\ref{#1}}

\def\secref#1{Section~\ref{#1}}
\def\tabref#1{Table~\ref{#1}}
\def\eqref#1{(\ref{#1})}

\def\ie{\emph{i.e.}\xspace}
\def\eg{\emph{e.g.}\xspace}

\newcommand{\argmin}[1]{\mathop{\arg\min}_{#1}\hspace{0.5em}}

\renewcommand{\bar}[1]{\overline{#1}}

\def\mathlet#1#2{\pgfmathparse{#2}\let#1\pgfmathresult}

\DeclareMathOperator{\IGM}{IGM}
\DeclareMathOperator{\IG}{IG}
\DeclareMathOperator{\RIG}{RIG}

\icmltitlerunning{Instance Separation Emerges from Inpainting}

\begin{document}

\newcommand{\pinpaint}{\ensuremath{p_{\theta}}\xspace}

\newcommand{\infogain}{\ensuremath{\IG}\xspace}
\newcommand{\relinfogain}{\ensuremath{\RIG}\xspace}
\newcommand{\IGmeasure}{\ensuremath{\IGM}\xspace}
\newcommand{\indloss}{\ensuremath{\mathcal{L}_{\text{ind}}}\xspace}
\newcommand{\indlosssmooth}{\ensuremath{\mathcal{L}_{\text{ind}}}\xspace}
\newcommand{\indlosseff}{\ensuremath{\mathcal{\tilde{L}}_{\text{ind}}}\xspace}
\newcommand{\iploss}{\ensuremath{\mathcal{L}_{\text{inpaint}}}\xspace}

\newcommand{\allpixels}{\ensuremath{\Omega}\xspace}

\newcommand{\gtaffs}{\ensuremath{A^*}\xspace}
\newcommand{\affs}{\ensuremath{A}\xspace}
\newcommand{\gtseg}{\ensuremath{S^*}\xspace}
\newcommand{\seg}{\ensuremath{S}\xspace}

\newcommand{\anysubset}{\ensuremath{N}\xspace}
\newcommand{\anyset}{\ensuremath{M}\xspace}
\newcommand{\softmask}{\ensuremath{m}}
\newcommand{\softmaskUP}{\ensuremath{\anyset_{\softmask > 0.5}}}
\newcommand{\softmaskDOWN}{\ensuremath{\anyset_{\softmask \le 0.5}}}
\newcommand{\optimset}{\ensuremath{V}\xspace}

\newcommand{\embedding}{\ensuremath{\phi}}

\newcommand{\todo}[1]{{\color{red}todo: #1}}

\newcommand\refeqq[2]%
{\stackrel{\scriptscriptstyle(\mkern-1.5mu\ref{#1},\ref{#2}\mkern-1.5mu)}{=}}

\newcommand\refle[1]%
{\stackrel{\scriptscriptstyle(\mkern-1.5mu\ref{#1}\mkern-1.5mu)}{\le}}

\newcommand{\specialcell}[2][c]{%
  \begin{tabular}[#1]{@{}c@{}}#2\end{tabular}}

\def\noisetovoid{\textsc{Noise2Void}\xspace}
\def\mutexws{\textsc{MutexWatershed}\xspace}
\def\pixtopix{\textsc{Pix2Pix}\xspace}

\def\ouraffs{\textsc{InpaintAff}\xspace}
\def\fgnet{\textsc{FgNet}\xspace}
\def\fgnetmid{\textsc{FgNet50}\xspace}
\def\affnet{\textsc{AffNet}\xspace}
\def\truefg{\textsc{TrueFg}\xspace}

\def\unet{\textsc{U-Net}\xspace}
\def\adam{\textsc{Adam}\xspace}

\def\hela{\textsc{HeLa}\xspace}
\def\panc{\textsc{Panc}\xspace}

\twocolumn[{
  \icmltitle{Instance Separation Emerges from Inpainting}
  \begin{icmlauthorlist}
\icmlauthor{Steffen Wolf}{iwr}
\icmlauthor{Fred A. Hamprecht}{iwr}
\icmlauthor{Jan Funke}{jan}
\end{icmlauthorlist}
\icmlaffiliation{iwr}{IWR/HCI, Heidelberg University, Germany}
\icmlaffiliation{jan}{HHMI Janelia, Ashburn, VA}
\icmlcorrespondingauthor{Steffen Wolf}{steffen.wolf@iwr.uni-heidelberg.de}
\icmlkeywords{Machine Learning, ICML, Cell Segmentation, Inpainting, Self-Supervised Learning}

  \vskip 0.3in
}]
\printAffiliationsAndNotice{}

\begin{abstract}
    
Deep neural networks trained to inpaint partially occluded images show a deep understanding of image composition and have even been shown to remove objects from images convincingly. In this work, we investigate how this implicit knowledge of image composition can be leveraged for fully self-supervised instance separation. We propose a measure for the independence of two image regions given a fully self-supervised inpainting network and separate objects by maximizing this independence. We evaluate our method on two microscopy image datasets and show that it reaches similar segmentation performance to fully supervised methods.
\end{abstract}

\newif\placeholders

\section{Motivation}
\begin{figure}[h!]
  \input{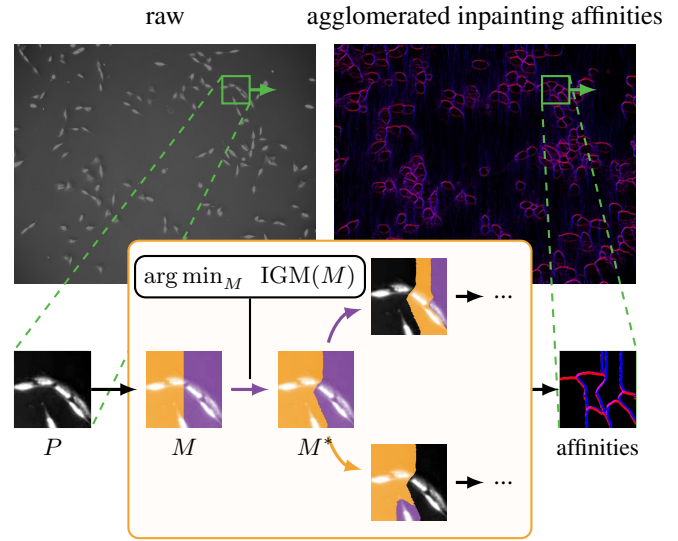}
  \caption{Extraction of instance separating affinities from an inpainting
  network.
  Given an image patch $P$, we optimize a set of pixels $\anyset$ (shown in purple)
  to minimize the \emph{information gain measure} \IGmeasure, which is based on
  the predictions of a probabilistic inpainting network (see
  \secref{sec:method:igm} and \figref{fig:method:igm} for details).
  This optimization ensures that pixels in $\anyset^*$ provide minimal
  information about the intensity values of pixels in the complement
  $\bar{\anyset}^*$ (shown in orange). We apply this procedure recursively to
  $\anyset^*$ and $\bar{\anyset}^*$ to obtain a hierarchical segmentation of
  the image patch from which we extract affinities (shown in blue/red for
  x-/y-direction, respectively). These affinities are computed and averaged
  over a set of sliding image patches (green box) to obtain the final affinity
  estimates.} \label{fig:introduction:overview}
\end{figure}

Recent inpainting neural networks demonstrate a remarkable ability to remove
distortions in natural images (\eg, text overlays, watermarks, or pixel-wise
independent noise) and are even able to entirely remove foreground objects
(\eg, a flagpole as demonstrated
\href{https://www.youtube.com/watch?v=gg0F5JjKmhA}{here}).
  Trained on large datsets, these networks learn the statistics that underlie
  images in a way that goes well beyond low level features.
  In this work, we aim to leverage those learnt statistics to distinguish
  individual objects in images from each other, without any form of
  supervision.

To intuitively understand how these statistics can be used, consider the following thought experiment. We train a high-capacity inpainting network on a very large corpus of
natural images and imagine the following scenario:
  Given the image of a busy street with a region in the center masked out to
  inpaint, such a network will be able to continue inpainting cars that are
  partially visible. If, however, the masked-out region is large enough to
  contain entire objects, the provided context will be uninformative about
  their location, shape, and texture and will therefore not be able to recover
  those objects.
  In other words, the success of predicting masked out objects depends on the
  information about those object contained in the surrounding context.

Here, we propose to exploit the predictability of image regions given partial
information to separate instances.
  We do so by maximizing the surprise of the inpainting network when trying to
  predict image content from one segment to another, or, equivalently, by
  minimizing the information gain between segments. This optimization can be
  carried out using only the implicit knowledge of inpainting networks about
  instances and thus gives rise to a self-supervised instance separation.

In particular, we define an information gain measure between image segments
that can be approximated efficiently given an inpainting network.
We show that minimizing this measure,
through a hierarchical optimization algorithm yields useful image decompositions.
We represent those decompositions by
\emph{affinities}, \ie, attractive or repulsive edges between pairs of
pixels, which we average over a set of image patches in a sliding window
fashion to obtain affinities for arbitrarily large images. An overview of this
method is shown in \figref{fig:introduction:overview}. The resulting affinities
require only minimal post-processing to obtain a segmentation.
We apply our method to the challenging problem of cell segmentation in
microscopy images, where we show that the unsupervised instance separation
finds non-trivial splits and is competitive with supervised methods.

\def\bsddatasetexamples{\href{https://www2.eecs.berkeley.edu/Research/Projects/CS/vision/bsds/BSDS300/html/dataset/images/color/25098.html}{fruits}, \href{https://www2.eecs.berkeley.edu/Research/Projects/CS/vision/bsds/BSDS300/html/dataset/images/color/22013.html}{fences}, or \href{https://www2.eecs.berkeley.edu/Research/Projects/CS/vision/bsds/BSDS300/html/dataset/images/color/124084.html}{flowers}}

\section{Self-Supervised Segmentation}

In general, self-supervised segmentation is an under-constrained problem.
What exactly constitutes a correct segmentation of an image depends not only
on the application context (\eg, segment all cells in a microscopy image),
but also on a subjective level of detail (\eg, segment nuclei and cell
membrane individually). Without constraining assumptions or instructions,
several different segmentations of the same image are plausible, leading to
an intrinsic ambiguity. This ambiguity can be prominently observed as the
inter-human variance for segmentation tasks where the concept of a segment is
not precisely defined (see, \eg, human generated segmentations of the BSD500
dataset for \bsddatasetexamples)\cite{amfm_pami2011}.

In the case of supervised image segmentation, this ambiguity is resolved by a
set of training object instances in the form of, \eg, affinities, labeled
images, bounding boxes, or polygons.
For self-supervised segmentation, on the other hand, assumptions about what
constitutes a segmentation have to fill in for the lack of training data.

Here, we propose to resolve this ambiguity by assuming that pixels of the same
instance are more predictable from each other than across instances.
We define the similarity between two pixels (and therefore the likelihood to
be part of the same instance) as the information gained about the value of one
pixel by observing the value of the other one.
In the following we will derive this similarity from a measure of inpainting
accuracy.

\subsection{Self-supervised Inpainting}

Let $x_i$ be a random variable representing the intensity of pixel $i \in
\allpixels$, and $x_\anyset$ with $\anyset \subseteq \allpixels$ a set of
random variables $\{ x_i \;|\; i \in \anyset \}$.
Probabilistic inpainting is equivalent to learning a parameterized function
$\pinpaint(x_i|x_\anyset)$, \ie, the conditional distribution over intensities
of pixel $i$, given known intensities of a partial observation
\anyset. The parameters $\theta$ of the distribution \pinpaint can be learned
by maximizing the likelihood of a measurement $x=x^*$, or equivalently by
minimizing the following negative log-likelihood:
\begin{equation}
	\iploss(\theta;\anyset) =
	\sum_{i\notin\anyset}
	-\log
	\pinpaint\left(x_i=x_i^*|x_\anyset=x^*_\anyset\right)
\end{equation}
It is worth noting that this loss formulation resembles the objective of
probabilistic \noisetovoid~\cite{krull2019probabilistic}, highlighting the
close connection between inpainting and denoising. In the next subsection, we
will derive a similar connection between inpainting (``predictability'') and
instance separation (``affinity'').

\subsection{Predictability is Affinity}
\label{sec:method:igm}

Our central assumption is that the intensity value of a pixel in an instance is conditionally independent of all pixels outside the instance. %

In other words, pixel values should be well predictable given the values of
other pixels in the same instance (high affinity). Conversely, pixel values
from other instances should provide \emph{no additional} information (low affinity).
More formally, let $\seg = \{ \seg_u \subseteq\allpixels \}$ be a segmentation
of \allpixels (\ie, $\bigcup_u \seg_u = \allpixels$ and $\forall u\neq v:
\seg_u\cap \seg_v = \varnothing$), and let $\seg(i)\subseteq\allpixels$ denote
the segment containing pixel $i$. We assume that for the true instance
segmentation $\gtseg$
\begin{equation}
  p(x_i|x_{\allpixels\setminus\{i\}}) = p(x_i|x_{\gtseg(i)\setminus\{i\}})
	\text{,}
	\label{eq:independence}
\end{equation}
or, equivalently, that there is \emph{no further} information gain provided by
\allpixels compared to $\gtseg(i)$ for estimating the value of $x_i$.
For general subsets $\anyset\subseteq\allpixels$, let $\infogain(i|\anyset)$
denote the additional information gained for estimating the value of $x_i$
when observing \allpixels compared to \anyset alone, \ie,
\begin{equation}
	\infogain(i | \anyset) =
	D_{\text{KL}} \Big(
	p(x_i | x_{\allpixels \setminus \{i\}})
	\;\Big|\Big|\;
	p({x_i} | x_{\anyset \setminus \{i\}})
	\Big)
	\text{,}
	\label{eq:infogain}
\end{equation}
where $D_\text{KL}$ denotes the Kullback-Leibler divergence. In the
following, we will use $\infogain(i|\anyset)$ as a measure of how much $x_i$
depends on values \emph{not} contained in \anyset.

Considering our assumption stated in \eqref{eq:independence}, a sensible
objective to recover a single segment of the true segmentation \gtseg would be
to minimize \eqref{eq:infogain} with respect to \anyset.
In practice, however, it would be unreasonable to assume that even for a
correct segment \anyset the information gain for pixels in this set from pixels
outside this set is exactly zero. In other words, dilating \anyset would
trivially decrease $\infogain(i|\anyset)$ until $\anyset = \allpixels$.
Therefore, instead of minimizing \eqref{eq:infogain} directly, we propose to
minimize a symmetric information gain measure. Let $\bar{\anyset} =
\allpixels\setminus\anyset$ be the complement of \anyset. Recall that
$\infogain(i|\anyset)$ measures the dependency of $x_i$ on values in
$\bar{\anyset}$. We introduce a \emph{relative} information gain that indicates
whether \anyset or $\bar{\anyset}$ provide more information about the value of
$x_i$:
\begin{equation}
	\relinfogain(i|\anyset) =
	\infogain(i|\anyset) -
	\infogain(i|\bar{\anyset})
	\text{.}
	\label{eq:relinfogain}
\end{equation}
The quality of a single segment \anyset can now be assessed by the following
symmetric information gain measure over all pixels $i$:
\begin{align}
	\IGmeasure(\anyset)
	=& \sum_{i \in \anyset}
	\relinfogain(i | \anyset)
	+ \sum_{i \in \bar{\anyset}}
	\relinfogain(i | \bar{\anyset})
  \\=&
    \sum_{i \in \anyset} \relinfogain(i | \anyset) -
    \sum_{i \in \bar{\anyset}} \relinfogain(i | \anyset)
	\label{eq:indloss}
	\text{.}
\end{align}

\subsection{Efficient Implementation}\label{sec:efficient_implementation}

In its current form, $\IGmeasure(\anyset)$ requires evaluation of
$\infogain(i|\anyset)$ for every pixel $i\in\allpixels$.
For each of these evaluations, $\pinpaint(x_i|\cdot)$ has to be computed
two times (conditioned on $\anyset$ and $\bar{\anyset}$), which is too inefficient
for a practical implementation.

To remedy this inefficiency, we make two approximations: 
First, we take advantage of convolutional neural network architectures that can inpaint an arbitrary set of pixels $\anysubset$ for
the same conditional~\cite{liu2018image}:
\begin{align}
	\prod_{i \in \anysubset} \: \pinpaint(x_i|\anyset \setminus \{i\}) \approx & \: \prod_{i \in \anysubset} \: \pinpaint(x_i| \anyset \setminus \anysubset)
\end{align}
A similar approximation technique was first proposed by \citet{krull2019noise2void} who argue that this approximation is error-free for convolutional neuronal networks, 
if all pixels in $\anysubset$ are spaced further apart than the field of view of the network. In our experiments, we find that even much denser subsets can be chosen without significant impact.
We will refer to $\RIG(i|\anyset)$ using this approximation as
$\RIG_N(i|\anyset)$ in the following.

Second, due to the limited field of view of the inpainting network%
, pixels far away from the conditional set have to be estimated via a constant prior and the relative information gain can therefore be computed without evaluating the neural network. Similarly, the complement conditional contains all pixels in the field of view. This is exactly the denoising setup of \noisetovoid ~\cite{krull2019noise2void}. Therefore, for low-noise-images one can directly approximate $\infogain(i| \Omega) \approx 0$ and otherwise apply the \noisetovoid as a preprocessing step to our method. Thus, $\RIG(i|\anyset) \approx \text{const}$ for pixels far away from the boundary between \anyset and $\bar{\anyset}$.

In conclusion, limiting the computation of $\IGM$ to a specified region $N$ close to the boundary combined with the approximate $\RIG_N$ leads to the following 
approximation:
\begin{align}
  \IGM_N(\anyset) &=
    \sum_{i\in {\anyset \cap N}}\hspace{-2mm}\RIG_N(i|\anyset) -
    \sum_{i\in {\bar{\anyset} \cap N}}\hspace{-2mm}\RIG_N(i|\anyset)
    \label{eq:method:igmn}
  \\
  &\approx \IGM(\anyset) + \text{const}
\end{align}

\subsection{Segmentation from Maximal Independent Regions}

\begin{figure}[t]
  \centerline{\input{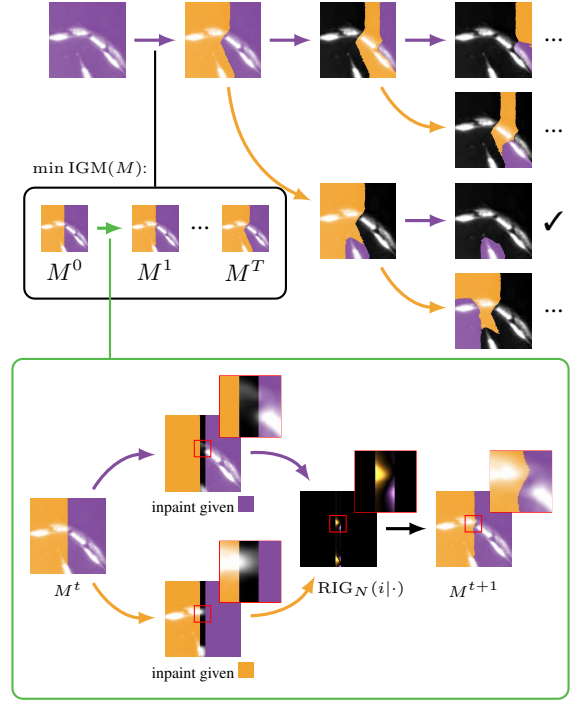}}
  \caption{Details of the hierarchical segmentation of an image patch from an
  inpainting network.
  Given an image patch (top left), we recursively find optimal splits (shown in
  orange and purple) by evolving a randomly chosen horizontal or vertical split
  over $T$ iterations (black box).
  For each step (illustrated in the green box), we evolve the boundary of the
  split by consulting a probabilistic inpainting network to predict the
  intensity of pixels in a region $N$ around the boundary, once given only the
  information contained in \anyset and once in its complement $\bar{\anyset}$.
  We then measure the relative information gain $\RIG_N$ in the inpainting
  region to determine which component (orange or purple) provided more
  information about the pixels in $N$ and reassign \anyset accordingly.
  }
  \label{fig:method:igm}
\end{figure}

Although the approximation $\IGM_N$ introduced above reduces the computational
burden of evaluating $\IGM$, finding an optimal mask
\begin{equation}
  \anyset^* = \argmin{\anyset}\IGM_N(\anyset)
\end{equation}
still remains intractable in general due to the combinatorial number of possible masks. We propose to solve this optimization
problem by following a greedy optimization strategy that generates a sequence
of masks $\anyset^t$ for $t \in \{0,\ldots,T\}$ such that
$\IGM_N(\anyset^{t+1}) \leq \IGM_N(\anyset^t)$, illustrated in \figref{fig:method:igm}.

To this end, we first separate \allpixels into two equally sized components
$\anyset^0$ and $\bar{\anyset^0}$ by randomly splitting them horizontally or
vertically. We then evolve the boundary of the split by evaluating
$\RIG_N(i|\anyset^t)$ for all pixels $i \in N$ in close proximity to the
current boundary between $\anyset^t$ and $\bar{\anyset^t}$.
The sign of $\RIG_N(i|\anyset^t)$ indicates whether $\anyset^t$ or
$\bar{\anyset^t}$ provide more information about the pixel $i$. We update
\anyset accordingly, \ie,
\begin{equation}
  \anyset^{t+1} = (\anyset^t \setminus N) \cup \{i \in N\;|\;\RIG_N(i|\anyset^t) > 0\}
  \text{,}
\end{equation}
which, by definition of \eqref{eq:method:igmn}, monotonically decreases
$\IGM_N$.

Finally, in order to obtain a decomposition of an image into arbitrarily many
maximally independent regions, we apply the minimization recursively to already
identified regions, \ie, we repeat the optimization procedure described above
on regions $\anyset^*$ and $\bar{\anyset^*}$, until either $\anyset^*$ or
$\bar{\anyset^*}$ are empty. Further implementation details on our neighborhood
selection can be found in the appendix.

In order to extract affinities for a full image we compute maximally
independent regions on a set of overlapping, sliding image patches and average
their affinities. This procedure is illustrated in
\figref{fig:introduction:overview}.

\section{Experiments on Microscopy Image Instance Segmentation}

\begin{figure*}[!ht]
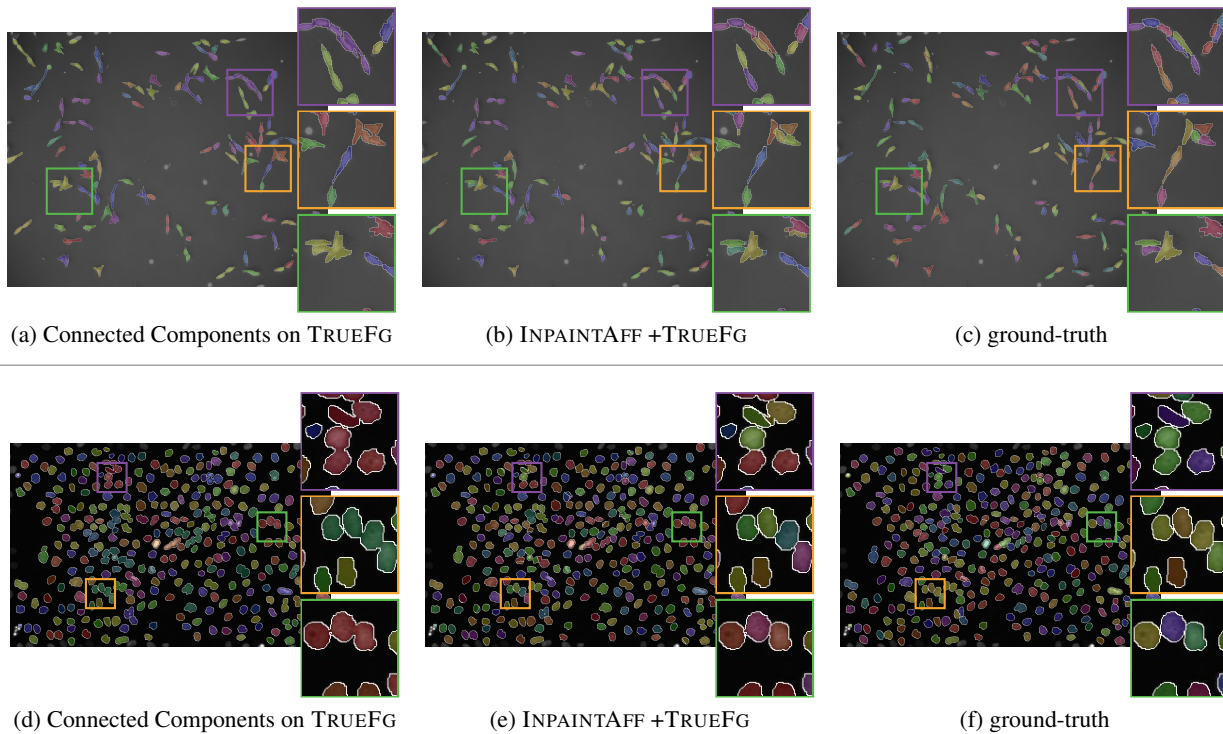

    \def\imgwidth{4.25cm}
    \def\zoomwidth{1.25cm}
    \centering
    \def\imgwidthpx{2816}
    \def\imgheightpx{2240}
    \def\zoomwidthpx{400}
    \def\zoomheightpx{400}
    \def\zoomAx{1940}
    \def\zoomAy{330}
    \def\zoomBx{2100}
    \def\zoomBy{1000}
    \def\zoomCx{350}
    \def\zoomCy{1200}
    \subfloat[Connected Components on \truefg]{
        \def\img{figures/results/PSC/Connected_components_on_gt}
        \input{figures/results/zoom.tikz}
    }
    \subfloat[\ouraffs+\truefg]{
        \def\img{figures/results/PSC/InpaintAff_TrueFG_seg_t01_122}
        \input{figures/results/zoom.tikz}
    }
    \subfloat[ground-truth]{
        \def\img{figures/results/PSC/gt_t01_122}
        \input{figures/results/zoom.tikz}
    }
    \vspace{2mm}
    {\color{gray}\hrule}
    \def\imgwidthpx{1100}
    \def\imgheightpx{700}
    \def\zoomwidthpx{100}
    \def\zoomheightpx{100}
    \def\zoomAx{300}
    \def\zoomAy{70}
    \def\zoomBx{260}
    \def\zoomBy{470}
    \def\zoomCx{850}
    \def\zoomCy{240}
    \subfloat[Connected Components on \truefg]{
        \def\img{figures/results/N2DL/Connected_components_on_gt}
        \input{figures/results/zoom.tikz}
    }
    \subfloat[\ouraffs+\truefg]{
        \def\img{figures/results/N2DL/InpaintAff_TrueFG_t02_75}
        \input{figures/results/zoom.tikz}
    }
    \subfloat[ground-truth]{
        \def\img{figures/results/N2DL/gt_t02_75}
        \input{figures/results/zoom.tikz}
    }
    \caption{Instance separation results assuming an accurate foreground
        detection \truefg on the \panc dataset (top row) and the \hela dataset
        (bottom row). A foreground detection alone is not sufficient to segment
        touching cells (a, d). \ouraffs extracted from an inpainting network find
    non-trivial splits between instances (b, e).}
    \label{fig:results:affvscc}
\end{figure*}

Instance separation is of particular importance for the identification and
tracking of individual cells in microscopy images, where cells frequently form
densely packed clusters and thus pose a challenging segmentation problem
\cite{ulman2017objective}.
In many cases, those cells are freely moving in a substrate and can thus be
considered as many independent instances of the same kind, which makes them
suitable for an inpainting based approach like the one we propose here and in
particular for the independence assumption we made in \eqref{eq:independence}.
In the following, we will refer to the affinities extracted using the proposed
method as \ouraffs.

\subsection{Cell Segmentation Benchmark Dataset}
\label{sec:experiments:dataset}

We evaluate \ouraffs on a subset of the ISBI Cell Segmentation Benchmark, which
includes a diverse set of 2D microscopy videos covering a wide range of cell
types and imaging quality.

In particular, we selected two datasets that contain cells of irregular shape
in close proximity for which instance separation is needed to obtain a
correct segmentation: (1) \hela contains cervical cancer cells expressing
H2b-GFP and (2) \panc contains pancreatic stem cells on a polystyrene substrate
(see~\figref{fig:experiments:affinities_and_segmentations} for samples and the
\href{http://celltrackingchallenge.net/2d-datasets/}{CTC website} for further
information about the datasets).

The \panc dataset arguably belongs to the more difficult datasets of the ISBI Cell
Segmentation Benchmark (reflected in the comparatively low test scores on the
challenge), which we attribute to two factors that are found in both \hela and \panc: First, they contain a large amount of touching cells with
little boundary evidence, which renders a mere foreground segmentation
ineffective for the detection of individual cells. Second, both datasets
contain only little labeled training data (815 instances\footnote{\hela has 571 additional instances, in partially labeled frames which can not trivially be used to train neural networks.} for \hela and 514 for \panc in fully labeled frames), which challenges fully supervised segmentation approaches.

\subsection{Results}
\label{sec:experiments:results}

\begin{figure*}
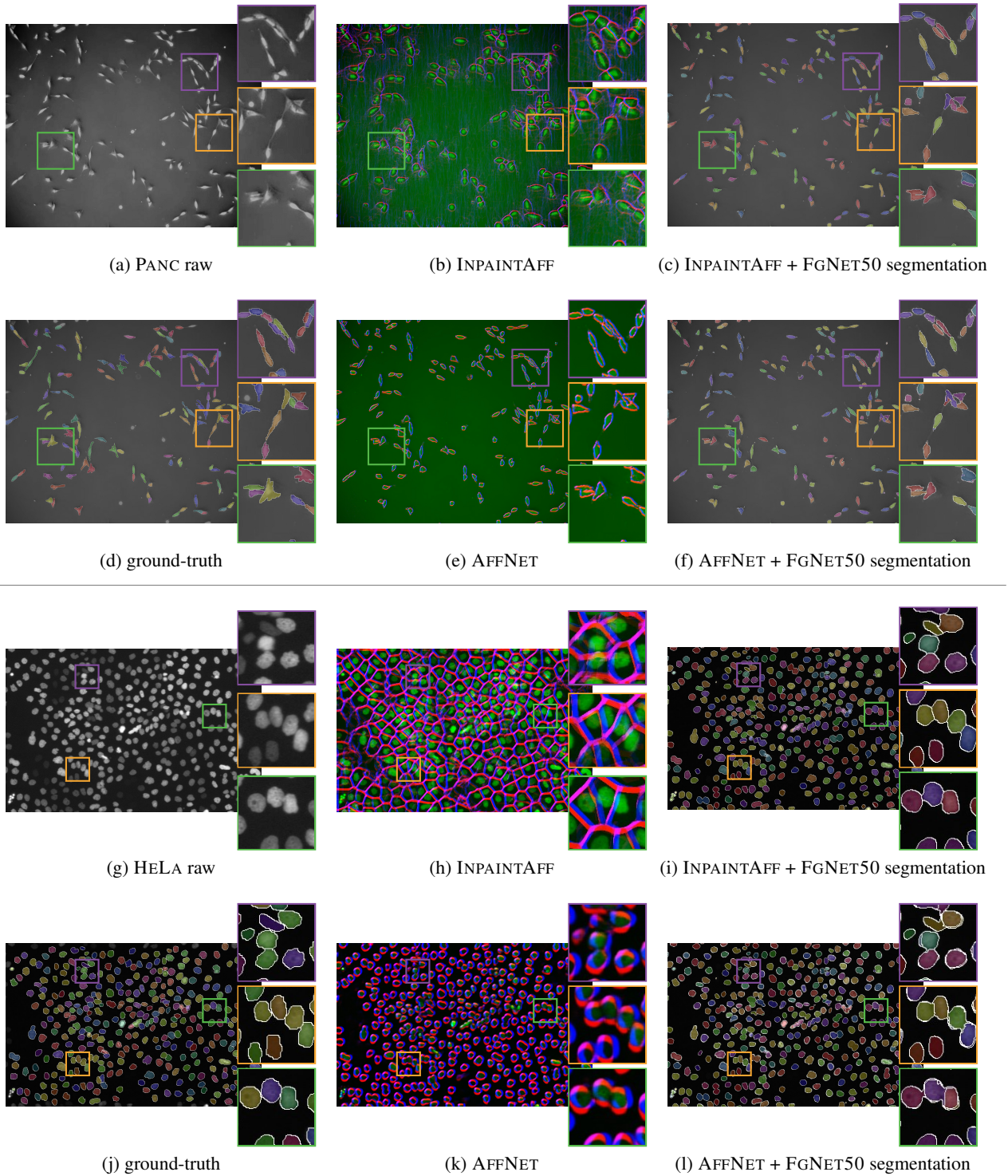

    \def\imgwidth{4.25cm}
    \def\zoomwidth{1.25cm}
    \centering
    \def\imgwidthpx{2816}
    \def\imgheightpx{2240}
    \def\zoomwidthpx{400}
    \def\zoomheightpx{400}
    \def\zoomAx{1940}
    \def\zoomAy{330}
    \def\zoomBx{2100}
    \def\zoomBy{1000}
    \def\zoomCx{350}
    \def\zoomCy{1200}
    \subfloat[\panc raw]{
        \def\img{figures/results/PSC/raw_t01_122}
        \input{figures/results/zoom.tikz}
    }
    \subfloat[\ouraffs]{
        \def\img{figures/results/PSC/raw_aff_inference_t01_122}
        \input{figures/results/zoom.tikz}
    }
    \subfloat[\ouraffs + \fgnetmid segmentation]{
        \def\img{figures/results/PSC/InpaintAff_FgNet_seg_t01_122_49instances}
        \input{figures/results/zoom.tikz}
        }\\
    \subfloat[ground-truth]{
        \def\img{figures/results/PSC/gt_t01_122}
        \input{figures/results/zoom.tikz}
    }
    \subfloat[\affnet]{
        \def\img{figures/results/PSC/SupervisedAff_t01_122}
        \input{figures/results/zoom.tikz}
    }
    \subfloat[\affnet + \fgnetmid segmentation]{
        \def\img{figures/results/PSC/SupervisedAff_FgNet_seg_t01_122}
        \input{figures/results/zoom.tikz}
        }\\
    \vspace{2mm}
    {\color{gray}\hrule}
    \def\imgwidthpx{1100}
    \def\imgheightpx{700}
    \def\zoomwidthpx{100}
    \def\zoomheightpx{100}
    \def\zoomAx{300}
    \def\zoomAy{70}
    \def\zoomBx{260}
    \def\zoomBy{470}
    \def\zoomCx{850}
    \def\zoomCy{240}
    \subfloat[\hela raw]{
        \def\img{figures/results/N2DL/raw_t02_75}
        \input{figures/results/zoom.tikz}
    }
    \subfloat[\ouraffs]{
        \def\img{figures/results/N2DL/raw_aff_inference_t02_75}
        \input{figures/results/zoom.tikz}
    }
    \subfloat[\ouraffs + \fgnetmid segmentation]{
        \def\img{figures/results/N2DL/InpaintAff_FgNet_t02_75_size52}
        \input{figures/results/zoom.tikz}
        }\\
    \subfloat[ground-truth]{
        \def\img{figures/results/N2DL/gt_t02_75}
        \input{figures/results/zoom.tikz}
    }
    \subfloat[\affnet]{
        \def\img{figures/results/N2DL/SupervisedAff_t02_75_size_52}
        \input{figures/results/zoom.tikz}
    }
    \subfloat[\affnet + \fgnetmid segmentation]{
        \def\img{figures/results/N2DL/SupervisedAff_FgNet_t02_75_size52}
        \input{figures/results/zoom.tikz}
    }
    \caption{Sample test images of \panc (top) and \hela (bottom). Affinities
    (middle column) are shown as blue/red for x-/y-direction, respectively.}
    \label{fig:experiments:affinities_and_segmentations}
\end{figure*}

As argued earlier, completely unsupervised segmentation is an under-constrained
problem. As such, \ouraffs alone is unlikely to give rise to a segmentation
capturing the intuition of a human annotator.
We recall that the main guiding principle for \ouraffs is predictability of
pixel intensities. Depending on the distribution of cells in images used to
train the inpainting network, this predictability might equally well apply to a
background region around each cell. This effect is visible in both datasets
(compare \figref{fig:experiments:affinities_and_segmentations}) and
demonstrates that the method is agnostic about the intensity of pixels and
merely clusters pixels that are mutually predictable.

Therefore, we investigate first how well \ouraffs \emph{separates} instances.
We then turn to the problem of instance segmentation, where we assume that at
least a small amount of ground-truth labels is available to capture the notion
of objects of interest---an assumption that arguably holds for any realistic
application in practice, where an accurate segmentation is required.

We report results using the ISBI Cell Segmentation Benchmark segmentation
accuracy (SEG score), a metric that is based on the Jaccard similarity index and
measures average IoU of all segments that overlap at least 50\% with the ground truth~(further details are given on the
\href{http://public.celltrackingchallenge.net/documents/SEG.pdf}{challenge
website}). The detection score is the percentage of matches that surpass a set IoU threshold.

\paragraph{Instance Separation}

\begin{table}[!ht]
    \centering
    \rowcolors{2}{gray!10!white}{gray!2!white}
    \begin{tabular}{llrr}
        \hline
        \rowcolor{gray!14!white}
        \multicolumn{2}{l}{Method} & \hela & \panc \\
        \hline 
        \multicolumn{2}{l}{Connected Components on \truefg} & 0.785 & 0.748 \\
        \multicolumn{2}{l}{\ouraffs + \truefg} & 0.858 & 0.914 \\ 
        \multicolumn{2}{l}{\ouraffs + \fgnetmid} & 0.766 & 0.666 \\ 
        \hline  
        HIT-CN$^*$   & MU-Lux-CZ$^*$ & 0.919 & 0.715 \\
        FR-Ro-GE$^*$ & CVUT-CZ$^*$   & 0.903 & 0.682 \\
        PURD-US$^*$  & HD-Hau-GE$^*$ & 0.902 & 0.665 \\
    \end{tabular}
    \caption{Segmentation scores assuming an accurate foreground detection
        \truefg and \fgnetmid (trained with 52/49 labeled instances for \hela/\panc).
        For reference, we include the official challenge scores of supervised methods
        on the same datasets (marked with a star), which have been trained on more
        labeled instances and evaluated on a different testing dataset then our
    method.}
    \label{tab:results:affvscc}
\end{table}

We investigate how well \ouraffs separates instances, assuming that an accurate
foreground segmentation is already available. For that, we use the ground-truth
segmentation provided in the datasets and convert it into a binary foreground
segmentation \truefg, while connecting all segments separated by a one pixel
wide gap.

As we show in \tabref{tab:results:affvscc} (and qualitatively in
\figref{fig:results:affvscc}), \truefg alone is not sufficient to achieve an
accurate instance segmentation, due to merges of cells in close proximity.
Separating those cells using \ouraffs, however, results in an almost perfect
instance segmentation, in the case of \panc even significantly exceeding the
scores of the best performing methods (albeit on different testing data and constrained to the ground-truth foreground). Those
results suggest that (1) \ouraffs is accurately separating instances, and (2) a
foreground segmentation is necessary and sufficient to constrain the boundaries
of found objects to obtain a competitive segmentation.

\paragraph{Instance Segmentation from Foreground Prediction}

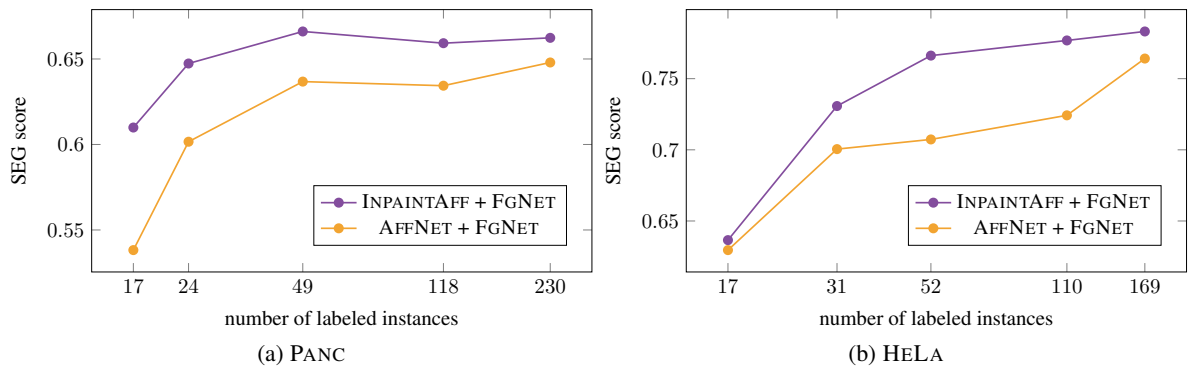
\begin{figure*}[tb]
    \centering
    \vspace{-2mm}
    \subfloat[\panc]{\def\plotfile{figures/plots/PSC.csv}\begin{tikzpicture}[scale=0.79,transform shape]
  \begin{semilogxaxis}[
    xlabel={number of labeled instances},
    ylabel={SEG score},
    legend style={at={(0.95,0.10)},anchor=south east},
    xtick=data,
    log ticks with fixed point,
    width=10cm,
    height=6cm,
    ]

    \addplot[funkey_color_1,thick,mark=*] table [
      x=labeled_instances,
      y=unsupervised_affinities]
        {\plotfile};
    \addlegendentry{\ouraffs + \fgnet}

    \addplot[funkey_color_2,thick,mark=*] table [
      x=labeled_instances,
      y=supervised_affinities]
        {\plotfile};
    \addlegendentry{\affnet + \fgnet}

  \end{semilogxaxis}
\end{tikzpicture}}
    \subfloat[\hela]{\def\plotfile{figures/plots/Hela.csv}\begin{tikzpicture}[scale=0.79,transform shape]
  \begin{semilogxaxis}[
    xlabel={number of labeled instances},
    ylabel={SEG score},
    legend style={at={(0.95,0.10)},anchor=south east},
    xtick=data,
    log ticks with fixed point,
    width=10cm,
    height=6cm,
    ]

    \addplot[funkey_color_1,thick,mark=*] table [
      x=labeled_instances,
      y=unsupervised_affinities]
        {\plotfile};
    \addlegendentry{\ouraffs + \fgnet}

    \addplot[funkey_color_2,thick,mark=*] table [
      x=labeled_instances,
      y=supervised_affinities]
        {\plotfile};
    \addlegendentry{\affnet + \fgnet}

  \end{semilogxaxis}
\end{tikzpicture}}
    \vspace{-2mm}
    \caption{Segmentation score on the test data of datasets \panc and \hela, for
        varying amounts of labeled instances used to train \fgnet and \affnet.}
    \label{fig:experiments:seg_scores}
\end{figure*}

\begin{figure}[h]
    \def\plotfile{figures/plots/iou_plots/PSC.csv}\begin{tikzpicture}[scale=0.8,transform shape]
  \begin{axis}[
    xlabel={IoU threshold},
    ylabel={detection accuracy},
    legend style={at={(0.02,0.02)},anchor=south west},
    xmin=0,
    xmax=1,
    width=10cm,
    height=8cm,
    ]

    \addplot[funkey_color_1,thick,only marks,mark=*] coordinates {(0.2, 0.2)};
    \addlegendentry[funkey_color_1]{\footnotesize \ouraffs + \fgnet}
    \addplot[funkey_color_1,dotted,thick,no marks] table [
      col sep=comma,
      x=IoUthreshold,
      y=size1]
        {\plotfile};
    \addlegendentry{17 labeled instances}
    \addplot[funkey_color_1,dashed,thick,no marks] table [
      col sep=comma,
      x=IoUthreshold,
      y=size4]
        {\plotfile};
    \addlegendentry{49 labeled instances}

    \addplot[funkey_color_2,thick,only marks,mark=*] coordinates {(0.2, 0.2)};
    \addlegendentry[funkey_color_2]{\footnotesize \affnet + \fgnet}
    \addplot[funkey_color_2,dotted,thick,no marks] table [
      col sep=comma,
      x=IoUthreshold,
      y=size1_supervised_affinities]
        {\plotfile};
    \addlegendentry{17 labeled instances}
    \addplot[funkey_color_2,dashed,thick,no marks] table [
      col sep=comma,
      x=IoUthreshold,
      y=size4_supervised_affinities]
        {\plotfile};
    \addlegendentry{49 labeled instances}
    \addplot[funkey_color_2,thick,no marks] table [
      col sep=comma,
      x=IoUthreshold,
      y=size16_supervised_affinities]
        {\plotfile};
    \addlegendentry{230 labeled instances}

    \addplot[funkey_color_3,thick,no marks] table [
      col sep=comma,
      x=IoUthreshold,
      y=unsupervised_affinities_affinities_bg_true]
        {\plotfile};
    \addlegendentry[funkey_color_3]{\ouraffs + \truefg}

    \addplot[gray,thick,no marks] table [
      col sep=comma,
      x=IoUthreshold,
      y=connected_components_bg_true]
        {\plotfile};
    \addlegendentry{CC on \truefg}

  \end{axis}
\end{tikzpicture}
    \caption{Detection accuracy over different IoU thresholds on \panc. Over a
        large range of IoU thresholds, \ouraffs in combination with a foreground
        network \fgnet trained on 49 labeled instances has a higher detection
        accuracy then the fully supervised method \affnet trained on 230 labeled instances.}
    \label{fig:experiments:det_scores}
\end{figure}
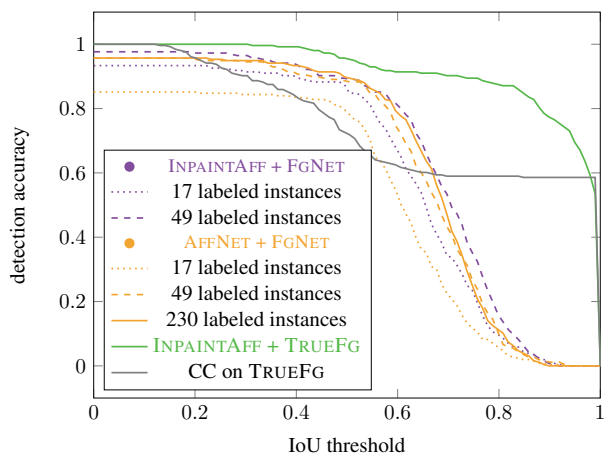

Since a foreground segmentation is crucial to capture the application specific
notion of what constitutes an object, we next investigate the segmentation
accuracy of our method when combined with a foreground prediction network
trained on few instances only, which we will refer to as \fgnet (details in
\secref{sec:experiments:details}).
We train \fgnet on varying amounts of labeled instances to predict a binary
foreground mask and use this prediction in combination with our \ouraffs to
obtain an instance segmentation.
As a baseline, we also train a second network \affnet to predict affinities
directly from the same labeled instances used to train the foreground network.

The segmentation scores for either approach on the test dataset are shown in
\figref{fig:experiments:seg_scores}, for varying amounts of labeled instances
used for training. Remarkably, \ouraffs consistently
outperform trained affinities in terms of the SEG score . This effect is most
visible in dataset \panc, where cells tend to cluster more compactly and the
separation of individual cells is therefore more challenging. In particular,
\ouraffs on this dataset in combination with \fgnet trained on as few as 24
labeled cells produce a segmentation that outperforms the fully supervised
\affnet using one order of magnitude more training data.
As shown in \figref{fig:experiments:det_scores}, this observation also holds in
terms of the detection score over varying IoU thresholds.
Furthermore, on the \panc datasets the obtained segmentation score using only around
50 labeled instances for the foreground prediction together with unsupervised
affinities is on par with the third leading submissions to the ISBI Cell Segmentation
Benchmark, which have been trained on 514 instances (albeit evaluated on
a different testing dataset then used here).

\subsection{Experiment Details}
\label{sec:experiments:details}

\paragraph{Training and Testing Split}

Since \ouraffs requires a considerable amount of computational resources (see
discussion in \secref{sec:limitations}) a direct evaluation on the CTC servers
on the official testing data is not possible. Therefore, we split the publicly
available data for each dataset into a train and testing dataset, each
containing one video of sparsely labeled cells. Further details can be found in
the appendix.

\paragraph{Model Architectures}

For the inpainting network underlying \ouraffs, we use a downscaled version of
the architecture proposed by~\citet{liu2018image}, \ie, a \unet architecture
with a depth of four resulting in five levels with 64, 128, 256, 512, and 512
feature maps, each. We train the network for 1M iterations using the \adam
optimizer and the loss proposed by~\citet{liu2018image} that is comprised of a perceptual, style, total variation and reconstruction loss.

\fgnet is a \pixtopix network \cite{zhu2017unpaired, isola2017image} with a depth
of six layers, containing 64 initial features maps, trained using \adam to
minimize a binary cross-entropy loss~\cite{kingma2014adam}. 

Since we use the \mutexws to post-process affinity predictions, we use the same
training procedure proposed by Wolf et al. \yrcite{wolf2018mutex} for \affnet~(\pixtopix architecture). In particular, we also use the S{\o}rensen-Dice coefficient \cite{dice1945measures,sorensen1948method} loss and the same affinity neighborhood (12 distances, up to 27 pixels).

\paragraph{Affinity-Based Segmentation}

We use the \mutexws to derive a segmentation from
affinities~\cite{wolf2018mutex}, where we introduce a single parameter $\alpha$
to control for over- and undersegmentation by multiplying all long range affinities (that are used to split) with $\alpha$. 
The optimal $\alpha$ for  each evaluated method was determined on the validation dataset.
For further details can be found in the appendix.

\section{Related Work}

While classical patch-based inpainting methods such as \cite{drori2003fragment,
sun2005image, barnes2009patchmatch} synthesize high quality images, they
fundamentally cannot make semantically aware decisions for intensity
predictions. Deep inpainting networks, on the other hand, trained on large
corpuses of data are known to develop an intrinsic understanding of images
\cite{larsson2017colorization}, which raises the question what aspects are
captured by these networks. The usefulness of these inpainting models for image
segmentation was shown by \citet{pathak2016context}, who demonstrate that
features extracted from a trained inpainting network capture appearance and
semantics of visual structures aiding in the pre-training of classification,
detection, and segmentation tasks.
Extending inpainting networks that directly minimize the reconstruction error
\cite{xie2012image, kohler2014mask} with texture and structure aware loss,
such as multi-scale neural patch synthesis \cite{yang2017high} or Structure-aware Appearance Flow \cite{ren2019structureflow}
leads to high-fidelity images and prediction and modeling of higher order relations

In parallel, specialized architectures and convolutions have been developed
that make it possible to realistically inpaint arbitrary masks
\cite{liu2018image, yu2019free}.

In this work, we use the network architecture and loss proposed by
\citet{liu2018image} which is designed to inpaint arbitrary masks and is
trained with an additional style component loss. Since we leverage the
network's learned distribution by measuring information gain between image
patches, we intentionally avoid networks trained with an additional GAN loss
\cite{nazeri2019edgeconnect, chen2018high, zeng2019learning}.
Although GANs produce extremely realistic looking images, they are prone to mode collapse that affects
our estimate of information gain.

More generally, inpainting falls under the broader category of unsupervised
prediction of left-out data, also known as \textit{self-supervised} learning
\cite{de1994learning}. This includes tasks such as image colorization
\cite{zhang2016colorful, larsson2016learning},  co-occurrence
\cite{isola2015learning}, predicting permutations \cite{anoop33deeppermnet}, and
denoising \cite{krull2019noise2void}. These methods are highly effective at
extracting robust features for further transfer learning \cite{zhang2017split}
and image embeddings \cite{trinh2019selfie} and can be considered a proxy task
for developing a semantic understanding \cite{larsson2017colorization}.

In some cases, the self-supervised task can be used as a free supervisory
signal that directly translates to classically supervised tasks. For example,
object tracking emerges from video colorization \cite{vondrick2018tracking}
(which inspired our title) or through obeying cycle-consistency in time
\cite{wang2019learning}. When provided with background images and images with
objects, \citet{ostyakov2018seigan} learn to segment by predicting masks and
paste patches from the object domain onto the background domain constrained by
an adversarial and a cycle consistency loss.

Our work uses the statistical properties of instances to derive a method for separating instances, 
which closely relates to other self-supervised segmentation approaches that utilize different properties
to identify objects. \citet{burgess2019monet} utilize compressibility, in a
compositional generative model, where image regions are reconstructed through a
low dimensional bottleneck. They show that their model is capable of discovering useful
decompositions of scenes by identifying segments that can be represented in a common format.
Another approach by \citet{chen2019unsupervised} learns to find masks
of objects by learning to replace the masked content content that corresponds
with altering the masked objects properties (e.g. altering the color of
flowers).

\section{Discussion}
\label{sec:limitations}

It remains an open question as to how far completely unsupervised segmentation
based on image statistics alone will find real world applications. As we
already observed on the segmentation of cells in microscopy images studied
here, an experimentalist's intention of what constitutes a good cell
segmentation does not necessarily match the clustering of pixels based on
information content. Only at least partially supervised methods with
application specific losses can ultimately produce predictions tailored to a
specific application, provided enough labeled training data is available.
We see the contribution of this work therefore primarily as an aid to
supervised methods, especially in scenarios in which labeled training data is
scarce. As our experiments demonstrate, \ouraffs allow practitioners to obtain
competitive segmentations from very few labeled instances. Given the high rate
and diversity of microscopy images acquired in the life sciences, self-supervised segmentation 
has the potential to significantly reduce the amount of human interaction
needed. Our work shows that in this domain the inherent
knowledge captured by inpainting networks provides 
competitive performance with very few labeled instances.

A limitation of the method proposed here is the runtime: \ouraffs requires
around 48h to process a 700x1100 image on a single GPU. Although inference can
be trivially parallelized, the current implementation might be prohibitively
slow for many applications. Increasing the efficiency of the inference by, \eg,
training networks directly on $\IGM$, will be subject of future work.

\bibliography{sefibib.bib}
\bibliographystyle{icml2020}

\appendix
\section{Neighborhood Selection and Inference:} 
As discussed, the updates of the equation~(11) can be limited to \anysubset, a set of pixels close to the boundary of  $\anyset$ and $\bar{\anyset}$.
Formally, let $\text{FOV}(i, d)$ be the set of all pixels closer to pixel $i$ than the max distance $d$. Then 
\begin{align}
\anysubset(\anyset, d) = \Big\{ i \in \allpixels \;\Big| \; &\text{FOV}(i, d) \cap \bar{\anyset} \neq \emptyset \text{ and } \\ 
&\text{FOV}(i, d) \cap \anyset \neq \emptyset\Big\}
\end{align}

Empirically, we find that decreasing $d$ over time aids the regions to converge.
In our experiments, we use a constant $d$ for the first half of the updates and then decrease it linearly. Additionally, we find that smoother boundaries can be achieved by interleaving updates with $d=1$ every second iteration and smoothing the reconstruction error over neighboring pixels. For the smoothing, we convolve the reconstruction error with gaussian kernels with $\sigma \in [0.1, 1, 5, 10]$ and add them to the pixel-wise reconstruction error.

\section{Train/Test Split of CTC}

Each dataset of the \href{http://celltrackingchallenge.net/2d-datasets/}{Cell Tracking Challenge} contains two training (labeled \textit{t01} and \textit{t02}) and two test videos. Since our inference method requires a considerable amount of computational resources, a direct evaluation on the CTC servers on the official testing data is not possible. Therefore, we split the publicly available data for each dataset into a train and testing dataset.

For the \textit{\panc~(PhC-C2DL-PSC)} dataset we train on frame 182 of video \textit{t02}, validate on frame 25 of \textit{t02} and test on frames $[98, 122]$ of video \textit{t01}. This uses all 4 available labeled frames of the dataset.

For the \textit{\hela~(Fluo-N2DL-HeLa)} dataset we train on frames $[13, 52]$ of video \textit{t01}, validate on frame 76 of \textit{t01} and test on all (even partially labeled)~frames $[23, 35, 36, 67, 75, 78, 79, 87]$ of video \textit{t02}.

The training sets with a reduced number of instances were generated by first, using a random subset of labeled frames and then cropping the training images spatially. We alternate between halving the image size in x and y-direction, taking away from both sides, thus keeping the center constant.

\section{Affinity-Based Segmentation}

We derive a segmentation from affinities $\text{aff}$ using the \mutexws 
on a XY-plane neighborhood graph with local attractive edges $[-1, 0], [ 0, -1]$ and sparse repulsive edges: $[-9, 0],  [0, -9],  [-9, -9],$ $[9, -9],  [-9, -4],  [-4, -9], [4, -9],  [9, -4],  [-27, 0],  [0, -27]$.
The graph weights for the local attractive edges are equivalent to the affinities, and the costs of the repulsive edges are the $\alpha$-weighted inverted affinities $\alpha (1 - \text{aff})$.

\end{document}